\documentclass[fullpaper,cameraready]{nldl}
\renewcommand{\DOI}{10.7557/18.6824}
\usepackage[utf8]{inputenc}
\usepackage{url}
\usepackage{graphicx}
\usepackage{authblk}

\usepackage{algorithm}
\usepackage{algorithmic}

\usepackage[square,numbers]{natbib}

\usepackage{hyperref}

\usepackage{tabularx}
\usepackage{amsmath}
\usepackage{tikz} % Illustrations
\usetikzlibrary{shapes,arrows,fit,positioning}

\tikzstyle{block} = [rectangle, draw,  
    text width=5em, text centered, rounded corners, minimum height=4em]
\tikzstyle{line} = [draw, -latex']
\tikzstyle{padding} = [inner sep=1em]

\usepackage{booktabs} % fancy tables
\usepackage{float}

\title{A~contrastive~learning~approach~for individual~re-identification~in~a~wild~fish~population}

\author[1]{{\O}rjan~Lang{\o}y~Olsen}
\author[2]{Tonje~Knutsen~S{\o}rdalen}
\author[1]{Morten~Goodwin}
\author[3]{Ketil~Malde}
\author[4]{Kristian~Muri~Knausg{\aa}rd\thanks{Corresponding kristianmk@ieee.org}}
\author[3]{Kim~Tallaksen~Halvorsen}
\affil[1]{Centre for Artificial Intelligence Research, University of Agder, Norway}
\affil[2]{Centre for Coastal Research, University of Agder, Norway}
\affil[3]{Institute of Marine Research, Ecosystem Acoustics Group, Bergen, Norway}
\affil[4]{Top Research Centre Mechatronics, University of Agder, Norway}

\date{\vspace{-5ex}}

\begin{document}
\nldlmaketitle

\begin{abstract}
In both terrestrial and marine ecology, physical tagging is a frequently used method to study population dynamics and behavior. However, such tagging techniques are increasingly being replaced by individual re-identification using image analysis.

This paper introduces a contrastive learning-based model for identifying individuals. The model uses the first parts of the Inception v3 network, supported by a projection head, and we use contrastive learning to find similar or dissimilar image pairs from a collection of uniform photographs. We apply this technique for corkwing wrasse, \textit{Symphodus melops}, an ecologically and commercially important fish species. Photos are taken during repeated catches of the same individuals from a wild population, where the intervals between individual sightings might range from a few days to several years.

Our model achieves a one-shot accuracy of 0.35, a 5-shot accuracy of 0.56, and a 100-shot accuracy of 0.88, on our dataset.

\end{abstract}

\section{Introduction\label{chap:intro}}

Physical tagging, using external or internal markings for individual identification, is a widely used method for monitoring terrestrial and aquatic animal populations. Information from resightings or recapture of the same individuals can be used to estimate population size, survival and movement patterns. However, most tagging methods are costly, intrusive, and labor-intensive. To our beneift, many animals have natural markings or morphological features that are unique to individuals that could be used for photo-identification and replace the need for physical tags \cite{weinstein2018computer,schneider2020similarity}. However, for ecologists, working with fish may mean keeping track of hundreds or potentially thousands of individuals in a population, which makes manual photo-identification challenging, if not impossible. For this reason, fully- or semi-automatic tools for re-identification of individuals would be immensely useful for ecologists.

Re-identification (re-ID) is different from normal classification in that it is a few-shot learning problem. Few-shot problems are characterised by having few samples per class, but there may be a large or indefinite number of classes. One way to solve such problems is a technique called metric learning, where data is transformed into embeddings of a lower dimension, that clusters points from the same class together. Classification can then be performed on the embeddings. Metric learning approaches have been proved to work well for re-identification of animal species \cite{schneider2019past}. A crucial advantage with metric learning approaches is that the network does not need to be retrained to be able to add new classes.

Contrastive learning is a technique that can be used to solve few-shot problems. Constrastive learning compares data and identifies whether they are similar or dissimilar. A siamese network \cite{bromley1993signature} is the most basic form and takes two inputs through the same network with the same shared weights and gets an embedding for both. During training, it tries to predict whether they are of the same class or not. A major advantage here is that it does not need to know which class an input belongs to, nor how many classes there are. Triplet networks \cite{hoffer2015deep} are an improvement to the siamese network with three inputs.

The goal of this work was to test the applicability of image based re-ID analysis for a commercially and ecologically important fish species, the corkwing wrasse (\textit{Symphodus melops}). The image dataset consists of standardized photos of captures and recaptures of individuals in a wild population, where the time between individual sightings spans from days to several years. The first step is to detect a fish in an image with an object detector, followed by a re-identification method. With high enough precision, computer vision re-ID has the potential to replace physical tagging for individual identification and may be applied in monitoring of survival rates, growth, movement, and population size, key knowledge for sustainable management and conservation \cite{schneider2019past} \cite{goodwin2022unlocking}.

\section{Related works}

Advancements in machine learning have produced powerful techniques for extracting ecologically important information from image and video data. For instance, machine learning have successfully been utilized to detect fish wounds \cite{gupta2022accurate}, count and categorize organisms in digital photos and real-time video \cite{lopez2020video}, \cite{knausgaard2022temperate}, identify species, \cite{gupta2022hierarchical}, and discover, and count creatures from digital images, \cite{goodwin2022unlocking}, and even quantify their behaviour \cite{Ditria2021}.

Some work on the topic of re-identification of fish has been conducted, but work on wild teleost fish are lacking. \citet{bruslund2020re} achieved an mAP of 99\% on Zebrafish using metric learning with 15 samples per class of 6 classes.
\citet{meidell2019fishnet} reports a true positive rate of 96\% on 225 thousand images of salmon divided between 715 individuals. 
\citet{li2022fish} achieved an accuracy of 92\% using 3412 images of 10 individuals using their novel FFRNet network.
These studies have in common that they were carried out in captivity and  are not using temporally independent observations. In other words, the individuals did not change morphology through growth, maturation, senescence, or similar biological processes.
\citet{moskvyak2019robust} used a metric learning approach on a dataset of 1730 images of 120 manta ray individuals and achieved an accuracy@1 of 62\% and an accuracy@10 of 97\%.
 
\section{Method}
 
\subsection{Data collection}

The study species, \textit{S. melops}, is a commercially and ecologically important species in coastal ecosystems in the Northeastern Atlantic \cite{Halvorsen2016}. This species have two distinct male morphs, colourful large males that build nest and care for the eggs, and smaller sneaker males, with a more brown coloration resembling the female morphology (brown and gray) \cite{Uglem2000}. The dataset was collected in Austevoll, western Norway, 2018-2021, by catching corkwing wrasse by fyke nets left in the sea overnight and marking all captured individuals with uniquely coded passive integrated transponder (PIT) tags (11 mm tags, RFID Solutions). The tags were implanted in the abdominal cavity of the fish, see full sampling description in \cite{Halvorsen2017} and \cite{Halvorsen2021}.

\begin{figure}[h]
\centering
\includegraphics[width=1.0\linewidth]{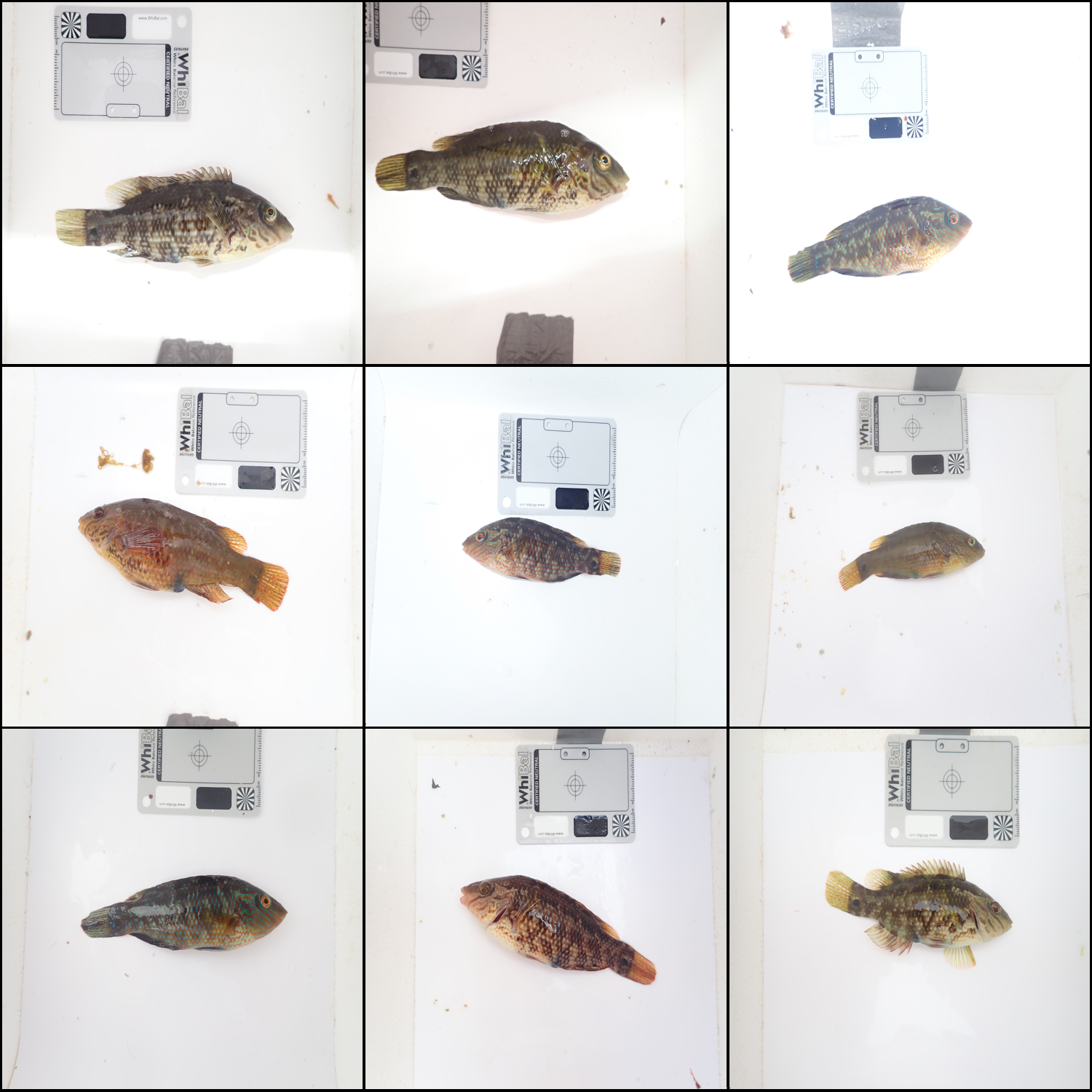}
\caption{Samples from the unprocessed dataset.}
\label{fig:dataset}
\end{figure}

This method enabled us to collect independent observations of each individual across time and for the dataset to encompass changes in the fish's morphology. At each capture, a few images were taken of the fish on both sides and the images were tagged with an id based on the RFID. The images are captured with the dorsal side of the fish facing up. After some filtration, a dataset that could be used for the task was compiled. The final dataset consists of 2113 images from 513 unique individuals. As an added statistic, the mean between the first and last capture-date of all the individuals is 230 days. Samples from the dataset can be seen in Figure~\ref{fig:dataset}.
 
\subsection{Individual re-identification}

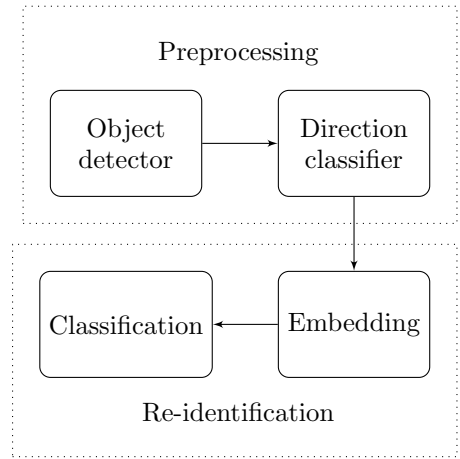
\begin{figure}
\centering
\begin{tikzpicture}[node distance = 2cm, auto]

\node [block] (obj) {Object detector};
\node [block, right of=obj, node distance=3cm] (side) {Direction classifier};
\node [block, below of=side, node distance=2.4cm] (embed) {Embedding};
\node [block, left of=embed, node distance=3cm, text width=] (class) {Classification};

\path [line] (obj) -- (side);
\path [line] (side) -- (embed);
\path [line] (embed) -- (class);

\node [above right=0.2cm and -2em of obj] (pre) {Preprocessing};
\node [fit=(obj) (side) (pre),draw,dotted,black,padding] {};

\node [below right=0.2cm and -3em of class] (main) {Re-identification};
\node [fit=(embed) (class) (main),draw,dotted,black,padding] {};

\end{tikzpicture}
\caption{\label{fig:pipeline}The network pipeline takes an image of a fish as input and outputs the id of the individual.}
\end{figure}

The re-identification system consists of a pipeline of different components, as illustrated in Figure~\ref{fig:pipeline}. The components fall into two categories, a preprocessing part and a re-identification part. As part of a preprocessing step in the pipeline, the system takes an image as input and feeds it to a object detection network to get an image crop, only containing the fish in the frame. Then a different network, the direction component, classifies whether the fish is facing right or left and passes this as metadata. For the re-identification part, the preprocessed data is fed to a contrastive learning network that learns to group embeddings for the same individual together and different apart. Classification can then be performed on the embeddings. By storing the embeddings of all previously observed individuals, re-identification can be achieved by nearest neighbor methods.

The object detector uses YOLOv5 \cite{glenn_jocher_2021_5563715} with an image size of 416x416, a batch size of 32 and is trained for 50 epochs. During training, the network was provided with manually annoted bounding boxes enclosing the fish.

The direction network is an Inception~v3~\cite{Szegedy_2016_CVPR} model with all its weights frozen. A global average pooling layer, a ReLU activated layer with 32 neurons, and a sigmoid activated output have been appended to the network. The dataset used for the training is the images cropped to only contain the head. The dataset is manually annotated with the direction.

The embedding network consists of a CNN model with a projection head. Its constituent parts were found experimentally. The CNN model is an Inception v3 model pre-trained on ImageNet, with the layers after the fourth concatenation layer (layer 46, or 132 if counting activation layers) removed. Appended at the end is a 2D global average pooling layer and a 128-dimensional linear projection that is normalized to the unit hypersphere. The network diagram is shown in Figure~\ref{fig:network}. The input size of the network is 224, and the images are resized accordingly before being fed into it. The network utilizes letter-boxing to maintain aspect-ratio. For the training of the embedding network, the dataset is split into a training and test set with a test set fraction of 0.3.

\begin{figure*}[ht]
\centering
\begin{tikzpicture}[
    -,
    >=stealth',
    auto,
    thick,
    node distance=0.5cm and 1cm,
    node/.style={rectangle,draw, minimum height=0.5cm, minimum width=0.25cm},
    legend/.style={rectangle,draw, minimum height=0.25cm, minimum width=0.25cm},
    conv/.style={fill=black!20!yellow},
    maxpool/.style={fill=black!20!red},
    avgpool/.style={fill=black!20!blue},
    concat/.style={fill=black!20!green},
    globavg/.style={fill=black!20!cyan},
    dense/.style={fill=black!20!orange},
    l2/.style={fill=black!20!pink},
]

\node[node, conv] (1) {};
\node[node, conv] (2) [right of=1] {};
\node[node, conv] (3) [right of=2] {};
\node[node, maxpool] (4) [right of=3] {};
\node[node, conv] (5) [right of=4] {};
\node[node, conv] (6) [right of=5] {};
\node[node, maxpool] (7) [right of=6] {};

\coordinate [right = 0.25cm of 7] (7f8) {};

\node[node, conv] (8b1) [above right=of 7, xshift=0.25cm] {};
\node[node, conv] (8b2) [right of=8b1] {};
\node[node, conv] (8b3) [right of=8b2] {};

\node[node, conv] (8a1) [above of=8b2, yshift=0.25cm] {};

\node[node, conv] (8c1) [below right=of 7] {};
\node[node, conv] (8c2) [right of=8c1] {};
\node[node, conv] (8c3) [right of=8c2] {};
\node[node, conv] (8c4) [right of=8c3] {};

\node[node, avgpool] (8d1) [below of=8c2, yshift=-0.25cm] {};
\node[node, conv] (8d2) [right of=8d1] {};

\node[node, concat] (9) [above right=of 8c4] {};

\coordinate [left = 0.25cm of 9] (8f9) {};
\coordinate [right = 0.25cm of 9] (9f10) {};

\node[node, conv] (10b1) [right of=9f10] {};
\node[node, conv] (10b2) [right of=10b1] {};
\node[node, conv] (10b3) [right of=10b2] {};

\node[node, conv] (10a1) [above of=10b2, yshift=0.25cm] {};

\node[node, maxpool] (10c1) [below of=10b2, yshift=-0.25cm] {};

\coordinate [right = 0.25cm of 10b3] (10f11) {};

\node[node, concat] (11) [right=0.25cm of 10f11] {};

\draw [dashed] (9.97,-1) -- (9.97,1);

\node[node, globavg] (12) [right of=11] {};
\node[node, dense] (13) [right of=12] {};
\node[node, l2] (14) [right of=13] {};

\path[every node/.style={font=\sffamily\small}]
    (1) edge node [right] {} (2)
    (2) edge node [right] {} (3)
    (3) edge node [right] {} (4)
    (4) edge node [right] {} (5)
    (5) edge node [right] {} (6)
    (6) edge node [right] {} (7)
    (7) edge node [right] {} (7f8)
    
    (7f8) edge[bend left=45] node [right] {} (8a1)
    (8a1) edge[bend left=45] node [right] {} (8f9)
    
    (7f8) edge[bend left=45] node [right] {} (8b1)
    (8b1) edge node [right] {} (8b2)
    (8b2) edge node [right] {} (8b3)
    (8b3) edge[bend left=45] node [right] {} (8f9)
    
    (7f8) edge[bend right=45] node [right] {} (8c1)
    (8c1) edge node [right] {} (8c2)
    (8c2) edge node [right] {} (8c3)
    (8c3) edge node [right] {} (8c4)
    (8c4) edge[bend right=45] node [right] {} (8f9)
    
    (7f8) edge[bend right=45] node [right] {} (8d1)
    (8d1) edge node [right] {} (8d2)
    (8d2) edge[bend right=45] node [right] {} (8f9)
    
    (8f9) edge node [right] {} (9)
    (9) edge node [right] {} (9f10)
    
    (9f10) edge[bend left=45] node [right] {} (10a1)
    (10a1) edge[bend left=45] node [right] {} (10f11)
    
    (9f10) edge node [right] {} (10b1)
    (10b1) edge node [right] {} (10b2)
    (10b2) edge node [right] {} (10b3)
    (10b3) edge node [right] {} (11)
    
    (9f10) edge[bend right=45] node [right] {} (10c1)
    (10c1) edge[bend right=45] node [right] {} (10f11)
    
    (11) edge node [right] {} (12)
    (12) edge node [right] {} (13)
    (13) edge node [right] {} (14)
    ;

\node [above right=0.4cm and 1cm of 8b3] {3x};
\node [fit=(8a1) (8d1) (9),draw,dotted,black,padding,fill=black,opacity=0.1,xshift=-0.6cm, minimum width=4cm] (block) {};

\matrix [below left] at (current bounding box.south east) {
  \node [legend,conv,label=right:Convolution] {}; &
  \node [legend,maxpool,label=right:MaxPool] {}; &
  \node [legend,avgpool,label=right:AvgPool] {}; \\
  \node [legend,concat,label=right:Concatenation] {}; &
  \node [legend,globavg,label=right:GlobalAvg2D] {}; &
  \node [legend,dense,label=right:Dense] {}; \\
  \node [legend,l2,label=right:L2 Normalization] {}; \\
};
    
\end{tikzpicture}
\caption{The embedding network utilizes the first part of Inception v3 \cite{Szegedy_2016_CVPR} with a custom projection head. The dashed line marks where the Inception part ends and the custom part starts. The grey part is repeated three times.}
\label{fig:network}
\end{figure*}

The training of the embedding network uses gradual unfreezing. The first 100 epochs have the layers before layer 29 frozen and a learning rate of 0.001, and the next 100 epochs have the layers before layer 18 frozen and a learning rate of 0.0001 for a total of 200 epochs. Layer 29 and layer 18 were selected because they are concatenation bottlenecks in the network architecture (green nodes in Figure~\ref{fig:network}). The loss function is online hard triplet mining with a margin of 1.0. Hard triplet mining is a technique where loss is only backpropagated for triplets where the negative is closer to the anchor than the positive. Thus, the use of online mining circumvents the need for three identical networks with shared weights. The training samples are randomly applied with image augmentations. A number between -20 and 20 is added to the hue and saturation. The image is rotated by a fraction between 0 and 0.1 in either direction, and a scale transformation between 0 and 0.1 is applied. The batch size used is 32.

Classification, and by extension re-identification, is done using a nearest neighbor approach. And in this case it is useful to define the training set as the support set and the test set as the query set. Nearest neighbor classification is non-parametric and does not need to be trained through optimization. The training step is simply to feed the support images through the embedding network and store the associated embeddings for the inference step. To classify an image, a query image is fed through the embedding network and then simply select the class of the nearest point of the query embedding to the support set embeddings. Source code for our implementation is available at GitHub\footnote{\url{https://github.com/orilan93/SiameseFish}}.

\subsection{Method for experiments}

The \textit{Symphodus melops} have a distinct high-contrast pattern in the head region (particularly on the operculum). For this reason, it would be useful to explore whether the network performs better on head crops than on crops of the whole body. The experiment is performed by training and evaluating the embedding network on images that are cropped to either part.

The system can also treat each side of the fish as different classes, and thus valuable information can be gained by doing inference on both, and then combining the results in an ensemble classifier. For this experiment, the dataset is split up into a left-sided section and a right-sided section such that there is a pairwise correspondance between the images. Two models are trained, where one is only for left-sided images and the other is only for right-sided images. The embeddings in the support set is sorted by the distance to the query image for each side. The predicted class is then the class which appears first when both sorted collections are taken into account.

An experiment to evaluate how well the system is able to distinguish between a re-sighted individual and an individual that has never been seen before was also conducted. A query embedding is considered a new individual if its distance is greater than a certain distance away from any support embedding. The query set was split into a test set and a validation set. A grid search was used to find a good distance threshold by maximizing the F1 score when evaluating the test set. The validation dataset for this experiment contains 317 samples.

\section{Results}

The metrics we use are accuracy@1, accuracy@5, and mAP@5. Accuracy@1 shows the correctness of the highest ranked category, i.e., the percentages of the highest predicted class are equal to the true class. Accuracy@5 shows the correctness of the five highest ranked categories, i.e., how many of the five highest classes contain the true class. mAP@5 similarly shows the precision of the five highest ranked categories, i.e., how many of the true categories are among the five highest ranked categories.

\subsection{Re-identification}

\begin{figure}[h]
\centering
\includegraphics[width=\linewidth]{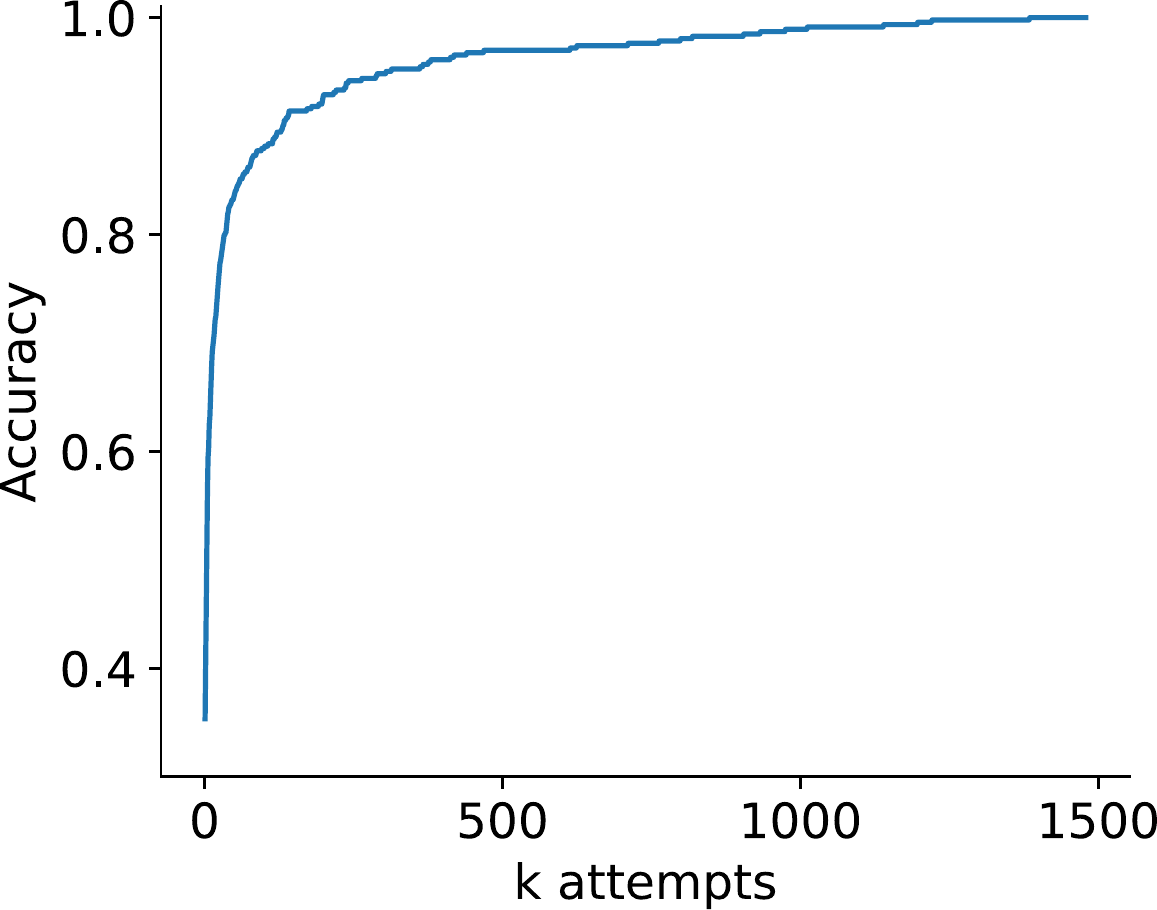}
\caption{The number of accumulated attempts (k) needed to attain a certain accuracy (accuracy@k).}
\label{fig:accuracy_curve}
\end{figure}

The re-identifcation system was evaluated against both the head and body crop datasets. Table~\ref{table:headvsbody} presents results from accuracy@1 and accuracy@5 and shows that the model performs best on the head crops. Figure~\ref{fig:accuracy_curve} shows how the model performs as the number of accumulated attempts increase.  This approach is essential in practice because, instead of having an unsorted catalog of images to go through, a professional biologist can go through a sorted catalog and expect to find the correct individual after inspecting the $k$ most promising images sorted based on the distance measure. The larger $k$ the higher accuracy, and as the number of attempts are approaching the number of images in the support set, the accuracy is approaching 100\%.

\begin{table}[h]
\centering
 \caption[]{Results for re-identification on head and body crops.}
 \begin{tabular}{l@{\hskip 0.19in} r@{\hskip 0.19in} r@{\hskip 0.19in} r} 
 \toprule
 Type & Accuracy@1 & Accuracy@5 & mAP@5 \\ %[0.5ex] 
 \midrule
Head & 0.3534 & 0.5647 & 0.4227 \\
 %\midrule
Body & 0.2043 & 0.3892 & 0.2690 \\
 \bottomrule
 \end{tabular}
 \label{table:headvsbody}
\end{table}

Table~\ref{table:distance} shows four random image samples from the dataset, together with the image the trained model predicts is the same individual and the ground truth. The classification rank and the distance in the embedding space are also shown.

\begin{table*}[t]
\centering
 \caption[]{The retrieval rank and euclidean distance between the embedding of a query image and a correct image.}
 \begin{tabular}{p{2.2cm} p{2.2cm} p{2.2cm} p{1.2cm} p{1.2cm}}
% \begin{tabular}{p{1.2cm} p{1.2cm} p{1.2cm} p{1cm} p{1cm}} 
 \toprule
 Query & Predicted & Ground truth & Rank & Distance \\ %[0.5ex] 
 \midrule
 \includegraphics[width=2.0cm,height=2.0cm,keepaspectratio=true]{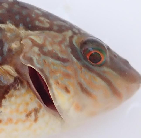} & \includegraphics[width=2.0cm,height=2.0cm,keepaspectratio=true]{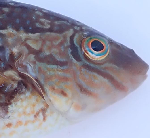} & \includegraphics[width=2.0cm,height=2.0cm,keepaspectratio=true]{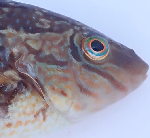} & 1 & 0.65 \\
 %\midrule
 \includegraphics[width=2.0cm,height=2.0cm,keepaspectratio=true]{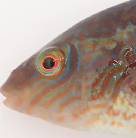} & \includegraphics[width=2.0cm,height=2.0cm,keepaspectratio=true]{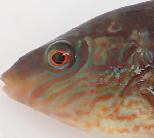} & \includegraphics[width=2.0cm,height=2.0cm,keepaspectratio=true]{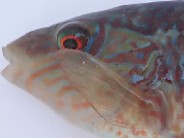} & 217 & 1.27 \\
 %\midrule
 \includegraphics[width=2.0cm,height=2.0cm,keepaspectratio=true]{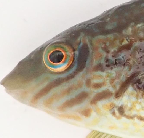} & \includegraphics[width=2.0cm,height=2.0cm,keepaspectratio=true]{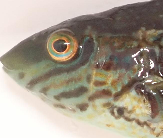} & \includegraphics[width=2.0cm,height=2.0cm,keepaspectratio=true]{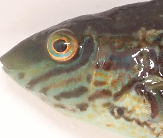} & 1 & 0.61 \\
 %\midrule
\includegraphics[width=2.0cm,height=2.0cm,keepaspectratio=true]{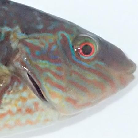} & \includegraphics[width=2.0cm,height=2.0cm,keepaspectratio=true]{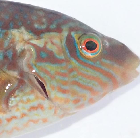} & \includegraphics[width=2.0cm,height=2.0cm,keepaspectratio=true]{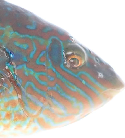} & 1012 & 1.46 \\
 \bottomrule
 \end{tabular}
 \label{table:distance}
\end{table*}

\begin{figure}[h]
\centering
\includegraphics[width=1.0\linewidth]{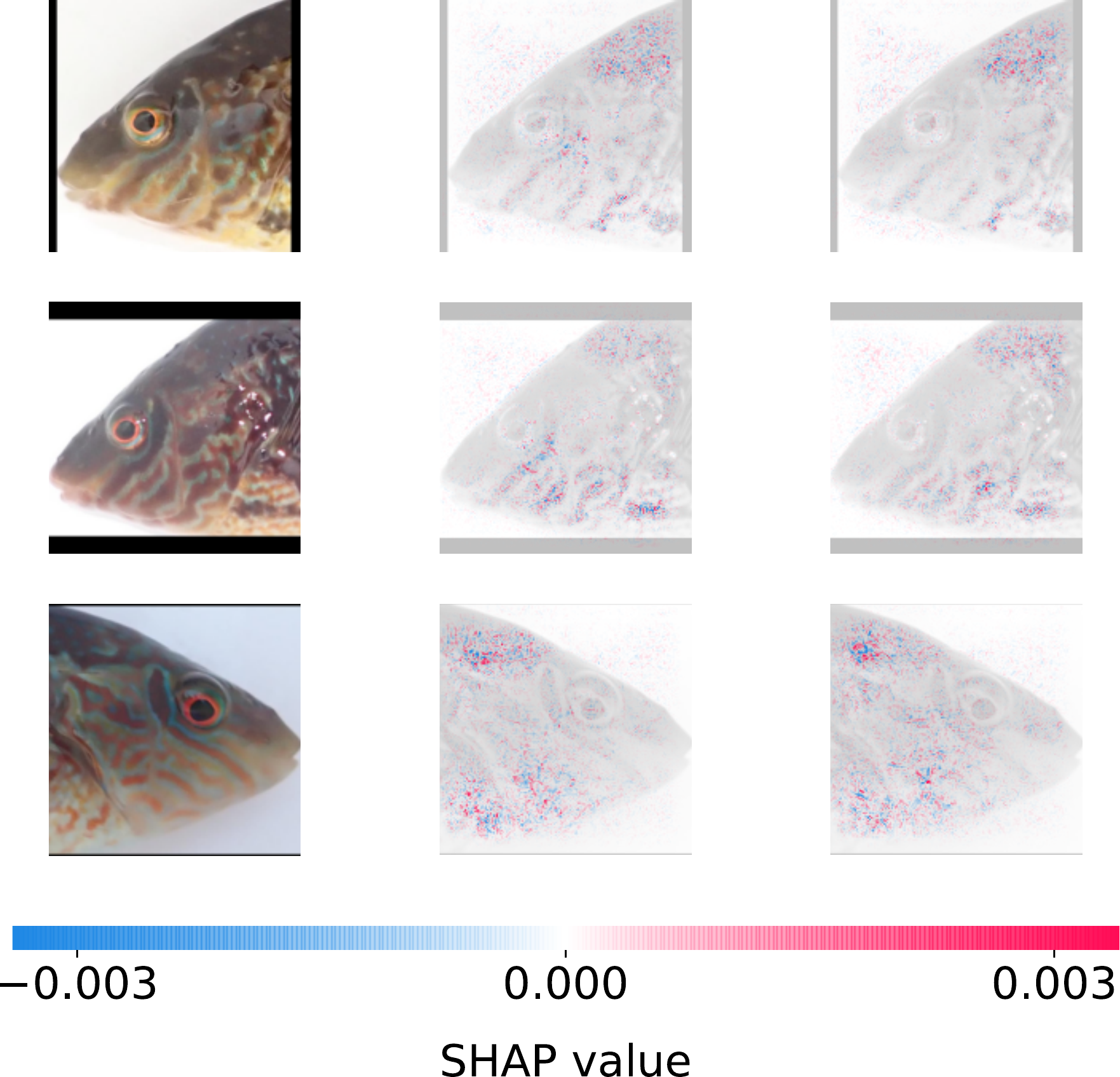}
\caption{SHAP plot showing which areas in the images that are most influential for the decisions of the model.}
\label{fig:shap}
\end{figure}

To gain insight into what the model focuses on when making its inferences, we present some test set samples and the accompanying SHAP plot \cite{lundberg2017unified} in Figure~\ref{fig:shap}. The colored area shows that the model is indeed picking up on the pattern of the fish.

\subsection{Ensemble classifier}
This experiment shows the results of training a new model for each side of the fish and then combining their respective classifications. Table~\ref{table:pairs} shows that this strategy can significantly increase performance. Note that the direction component, that is required for the ensemble classifier, yielded an accuracy@1 of 99.38\% on the validation set using the head cropped dataset.

\begin{table}[h]
\centering
 \caption[]{Ensemble classifier results.}
 \begin{tabular}{l@{\hskip 0.14in} r@{\hskip 0.14in} r@{\hskip 0.15in} r} 
 \toprule
 Type & Accuracy@1 & Accuracy@5 & mAP@5 \\ %[0.5ex] 
 \midrule
Left & 0.3568 & 0.5463 & 0.4243 \\
 %\midrule
Right & 0.4097 & 0.5595 & 0.4623 \\
 %\midrule
As pair & 0.5286 & 0.7533 & 0.6140 \\
 \bottomrule
 \end{tabular}
 \label{table:pairs}
\end{table}

\subsection{New observations}
As our previous experiments have shown, re-identification works relatively well. We aim at using this model for distinguishing new individuals from earlier observed individuals.
To identify new individuals with the model, an embedding distance threshold needs to be decided. Note that this relates to the distance metric in Table~\ref{table:distance}. Using grid search, we found a threshold of 0.820 to yield the best performance score on the validation set. The system predicted 95 individuals as new sightings and got a 62.78\% accuracy@1 at this task.

\section{Discussion and conclusion}

\begin{table}[h!]
\centering
 \caption[]{Summary of results.}
 \begin{tabular}{l@{\hskip 0.22in} r@{\hskip 0.22in} l} 
 \toprule
 Experiment & Result & Metric \\ %[0.5ex] 
 \midrule
Re-identification & 0.3534 & Accuracy@1 \\
 %\midrule
Object detector & 0.9951 & mAP@0.5 \\
 %\midrule
Direction classifier & 0.9937 & Accuracy@1 \\
 %\midrule
Ensemble classifier & 0.5286 & Accuracy@1  \\
 %\midrule
New observations & 0.6278 & Accuracy@1  \\
 \bottomrule
 \end{tabular}
 \label{table:summary}
\end{table}

Our experiments, summarized in Table~\ref{table:summary}, indicate that the system performs better on the head crops of the fish than on the whole body. This is likely because the pattern on the head is most distinct and thus an important feature, and this will appear at a higher resolution for the algorithm when resizing for the network input size. However, the drawback here is that the network is exposed to less information available in the data.

By utilizing the existing system in a new way by training separate models for each side of the fish, one can make an ensemble classifier. This method was tested and gained a considerable improvement from 35\% to 53\% accuracy. This shows how important it is to use all the information available to make good predictions.

The accuracy of this system is not high enough for a fully automated system with humans out-of-the-loop, which is required to replace the need for physical tags in ecological studies. However, we believe that continued collection of data can produce a dataset that is more temporally balanced to enable the model to account for the growth and ageing of the individuals.

Automatization can produce great benefits and is increasingly being adopted by many industries, and the field of ecology should be no different. A successful Re-ID algorithm with high precision can provide a new method with improved fish welfare, while also being cheaper (only a camera needed) and potentially more accurate (no tag loss). In the future, we envision that re-ID can be applied directly on live streams from under-water video cameras, removing the need for capture and handling fish altogether. This would be a revolutionary method that can drastically change how we can collect key information for sustainable conservation and management of fish and other animals.

\section*{Acknowledgements}
We thank Torkel Larsen, Anne Berit Skiftesvik, Ovin Holm, Ylva Vik, Nicolai Aasen, Ben Ellis, Vegard Omestad Berntsen, and Steve Shema for assistance in collecting the photos and capture-recapture data in the field. This study received funding from Centre for Artificial Intelligence Research (CAIR), Centre for Coastal Research (CCR), and Top Research Centre Mechatronics (TRCM) at University of Agder, the Institute of Marine Research (project 15638-01), and the Research Council of Norway (CoastVision, project number 325862, and CreateView, project number 309784).

%
% ---- Bibliography ----
%
\bibliographystyle{abbrvnat} % From NLDL2022 template
%\bibliography{references}

\end{document}